\documentclass[10pt,twocolumn,letterpaper]{article}

\usepackage{cvpr}
\usepackage{times}
\usepackage{epsfig}
\usepackage{epstopdf}
\usepackage{graphicx}
\usepackage{subfig}
\usepackage{float}
\usepackage[]{algorithm2e}
\usepackage{amsmath}
\usepackage{amssymb}
\DeclareMathOperator*{\argmax}{arg\,max}

\newcommand{\bD}[0]{\mathbf{D}}
\newcommand{\bE}[0]{\mathit{E}}
\newcommand{\bG}[0]{\mathcal{G}}
\newcommand{\bT}[0]{\mathcal{T}}
\newcommand{\bI}[0]{\mathbf{I}}
\newcommand{\bL}[0]{\mathit{L}}
\newcommand{\bP}[0]{\mathbf{P}}
\newcommand{\bS}[0]{\mathbf{S}}
\newcommand{\bW}[0]{\mathbf{W}}

\newcommand{\bx}[0]{\mathbf{x}}

\newcommand{\PPS}[0]{\textbf{PPS}}
\newcommand{\DPS}[0]{\textbf{DPPS}}
\newcommand{\RS}[0]{\textbf{RS}}
\newcommand{\US}[0]{\textbf{US}}
\newcommand{\QBC}[0]{\textbf{QBC}}

\newcommand{\comment}[1]{}

\cvprfinalcopy 

\def\cvprPaperID{1891} 

\ifcvprfinal\fi
\begin{document}

\title{Active Learning for Delineation of Curvilinear Structures}

\author{Agata Mosinska\\
EPFL\\
\and
Raphael Sznitman\\
University of Bern\\
\and
Przemys\l{}aw G\l{}owacki\\
EPFL\\
\and
Pascal Fua\\
EPFL\\
\and
{\tt\small  \{agata.mosinska, przemyslaw.glowacki, pascal.fua\}@epfl.ch raphael.sznitman@artorg.unibe.ch} 
}
\maketitle

\begin{abstract}
Many recent delineation techniques owe  much of their increased effectiveness to
path classification  algorithms that  make it possible to distinguish  promising paths
from others.   The downside of this  development is that they  require annotated
training data, which is tedious to produce.

In this paper,  we propose an Active Learning approach  that considerably speeds
up  the annotation  process. Unlike  standard ones,  it takes  advantage of  the
specificities of the delineation problem. It  operates on a graph and can reduce
the training  set size  by up  to 80\%  without compromising  the reconstruction
quality.

We  will  show  that  our  approach outperforms  conventional  ones  on  various
biomedical  and  natural  image  datasets,  thus  showing  that  it  is  broadly
applicable.
\end{abstract}


\section{Introduction}

Complex curvilinear structures are widespread in nature. They range in size from
solar filaments  as seen through telescopes  to road networks in  aerial images,
blood vessels in  medical imagery, and neural structures  in micrographs.  These
very  diverse structures  have different  appearances and  it has  recently been
shown that training classifiers to assess whether  an image path is likely to be
a  structure of  interest  is  key to  improving  the  performance of  automated
delineation
algorithms~\cite{Turetken12,Turetken13a,Breitenreicher13,Santamaria-Pang15,Montoya-Zegarra14,Wegner15}.

\begin{figure}[t!]
\centering
\subfloat[]{\includegraphics[height=0.23\textwidth,natwidth=512,natheight=512]{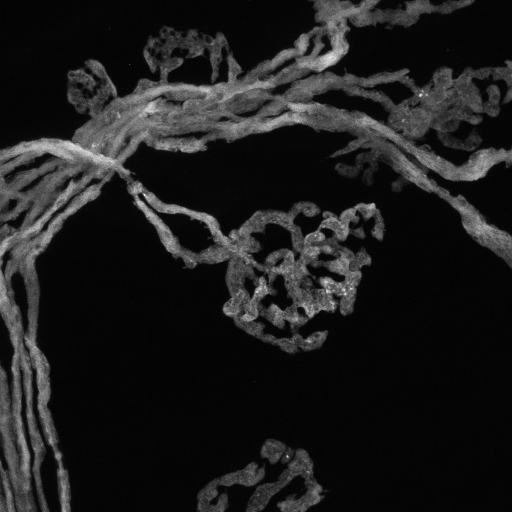}}\hspace{0.5em}
\subfloat[]{\includegraphics[height=0.23\textwidth,natwidth=512,natheight=512]{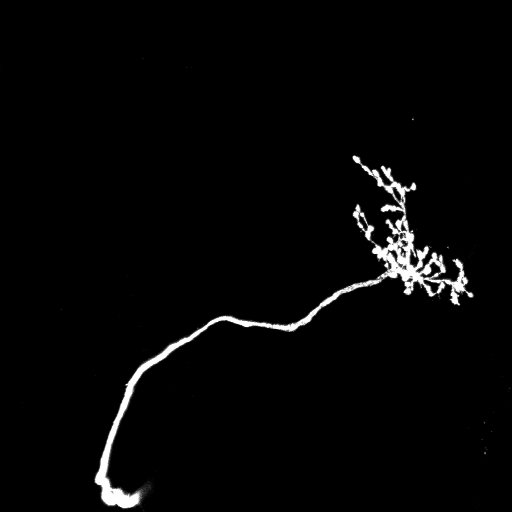}}

\caption{Images of two different neural structures obtained using confocal microscopy. The enormous variety of curvilinear structures requires problem-specific training datasets even in case of the same modality.}
\label{fig:differentNeurons}
\end{figure}

However, while such Machine-Learning based  algorithms are effective, they still
require significant  amounts of  manual annotation  for training  purposes.  For
everyday  scenes, this  can be  done by  crowd-sourcing~\cite{Long13,Lin14}.  In
more  specialized areas  such  as neuroscience  or medicine,  this  is impractical because only  experts whose  time  is scarce  and precious  can do  this
reliably. This problem is particularly acute  when dealing with 3D image stacks,
which  are much  more difficult  to  interact with  than regular  2D images  and
require  special expertise.  It  is further  compounded by  the  fact that  data
preparation processes tend  to be complicated and not  easily repeatable leading
the curvilinear structures to exhibit very different appearances
as  shown  in  Fig.~\ref{fig:differentNeurons}.  This means  that  a  classifier
trained on one  acquisition will not perform  very well on a new  one, even when
using the same modality.

In this  paper, we propose  an Active Learning  (AL) approach that  exploits the
specificities  of  delineation algorithms  to massively  reduce  the  effort  and
drudgery involved  in collecting  sufficient amounts of  training data.   At the
heart of all  AL methods is a query  mechanism that enables the system  to ask a
user to label a few well chosen data samples, which it has selected based
on  how informative  the answers  are  likely to  be.  AL  has been  successfully
deployed   in   areas   such    as   Natural   Language   Processing~\cite{Tong02},   Computer
Vision~\cite{Kapoor07},                                   and
Bioinformatics~\cite{LiuB04}.   While it  has made  it possible  to
train classifiers  with less of human  intervention, none of the algorithms can exploit the fact that, for delineation purposes,
the  paths to  be annotated  form a  graph in  which neighborhood  and geometric
relationships can and should be considered.

In our  approach, we explicitly  use these relationships to  derive multi-sample
entropy  estimates, which  are  better surrogates  of  informativeness than  the
entropy of  individual samples that  is typically used~\cite{Lewis94}. As  a result,
our queries focus more effectively on ambiguous regions in image space, that is,
those at the boundary between positive and negative examples.

To avoid  having to retrain the  system after each individual  query and further
increase efficiency, we  also integrate into our approach a  batch strategy that
lets the  system ask  the user  several  questions  simultaneously.   It
incorporates {\it  density} measures  that ensure that  the batches  are diverse in features,
representative of the delineation problem at hand and located near each other in
the images so as to facilitate  the interaction.  This is particularly important
in 3D volumes where scrolling from one  region to another far away is cumbersome
and potentially confusing for the user.

In short,  our contribution is an  AL approach that  is tailored for
the  delineation   of  complex  linear   structures.   In  that  sense,   it  is
specialized. However, it is also generic in  the sense that it can handle a wide
variety  of  different structures. We will show
that it  outperforms more  traditional approaches  on both  2- and  3-D datasets
representing different kinds of linear structures, that is, roads, blood vessels, and neural structures.

In the reminder of this paper, we first review existing AL techniques applicable
to  our  problem and  discuss  their  limitations  for  this purpose.   We  then
introduce  our approach  and show  how  we combine  information propagation  and
density measures  to streamline the  annotation process.  Finally,  we compare
the performance  of our  approach against conventional  techniques.


\section{Related Work}
\label{sec:related}

AL is predicated on the idea that, given even very small amount of
annotated data, the learning algorithm  can actively choose additional instances
that would be most profitable to  label next.  Starting from a small randomly
chosen and  manually annotated  set of  samples, iterating this  process can
drastically reduce  the need for  further human  annotation since only  the most
informative ones  are considered.   This has  been demonstrated  in applications
ranging from  Natural Language Processing  to Bioinformatics in  which unlabeled
data       is       readily        available       but       annotation       is
expensive~\cite{Settles11}.

All such AL  methods require a criterion for sample  selection. The most popular
one  is {\it  uncertainty}, usually  defined  as proximity  to the  classifier's
decision  boundary. When  the  classifier  is probabilistic,  this can  be
evaluated   in   terms   of   entropy~\cite{Settles08b}. In practice, Uncertainty Sampling can be incorporated into
most         supervised         learning         methods         such         as
SVMs~\cite{Tong02}, Boosting~\cite{Huang07} and Neural Networks~\cite{Cohn96}.

Another family of AL  algorithms called {\it query-by-committee}~\cite{Dagan95} uses
different automated ``experts''  to assign potentially different  labels to each
sample. Those  for which the disagreement is the greatest are prime  candidates for
manual annotation.

Most practical  AL algorithms  allow the  human to  annotate batches  of samples
before  retraining  the classifier.   This  spares  the need to wait  for 
potentially lengthy computations to finish between each intervention.
However, Uncertainty Sampling as described above can easily end up
querying outliers and in batch mode - redundant instances, which is inefficient. This is usually
addressed by considering not only the information gain potentially delivered
by  labelling  each individual  sample,  but  also  the representativeness of  each
batch, which is accomplished by \textit{density-based} methods. In~\cite{Settles08b}, Settles and Carven introduce a information density-weighted framework, which favours samples that are not only uncertain but also representative of the underlying distribution. The main problem associated with this approach is finding the weighting of the two terms. Li and Guo~\cite{Li2013} propose choosing a weight at each iteration that would minimise the future generalisation error. This approach is however computationally expensive, as it requires recomputing the underlying model many times and may additionally lead to overfitting.     

Most of  the methods discussed above  originate from fields other  than that of Computer
Vision.   They rarely  exploit  the  contextual or  spatial  relations that  are
prevalent      in     images      except     for      a     few      cases.
In~\cite{Siddiquie10} contextual  image properties are used  to find
the image regions  that would convey the most information  about other uncertain
areas       with      which       they      are       contextually      related. In~\cite{Aodha2014} a perplexity graph modelling similarities between images enables efficient hierarchical subquery evaluation.  In video segmentation application~\cite{Fathi11},  the obtained labels are propagated in a semi-supervised manner on a graph consisting of spatial, temporal and  prior edges. Then, the most uncertain frame is selected for the next annotation. We will show that propagating information \textit{after} preliminary classification and computing uncertainty only after this is advantageous over estimating informativeness based only on the result of classifier.  The AL approach to segmenting CT scans
of~\cite{Iglesias11} incorporates context  in  terms  of generative  anatomy
models.   The notion  of geometric  uncertainty for  segmentation is  introduced
in~\cite{Konyushkova15}.    Like  our   algorithm,  it   relies  on   exchanging
probability values between neighbours, but  does not account for dataset diversity.


\section{Active Learning for Delineation}
\label{sec:delin}

Graph-based  network  reconstruction  algorithms have  recently  shown  superior
performance compared to methods based on segmentation. They not only recover the
geometry of the problem, but also  the correct connectivity, which is crucial in
applications                               such                               as
neuroscience~\cite{Santamaria-Pang15,Wegner15,Montoya-Zegarra14,Turetken12,Turetken13a,neher2015}. They
largely owe their performance to supervised Machine Learning techniques
  that allow them to recognize promising linear paths.

These methods usually start by  computing a tubularity measure, which quantifies
the likelihood that a tubular structure  exists at given image location. Next, a
set                 of                subsampled                 high-tubularity
superpixels~\cite{Santamaria-Pang15,Wegner15,Montoya-Zegarra14}   or   longer
paths~\cite{Turetken12,Turetken13a,Breitenreicher13,neher2015}   are  extracted.
Each  of them  can be  considered  as an  edge $e_i$  belonging to  overcomplete
spatial  graph $\bG$  and characterized  by a  feature vector  $\bx_i$. Given two possible class labels $\mathit{(y_i = 1)}$ and $\mathit{(y_i = 0)}$  , a
discriminative  classifier assigns  to each  edge $\mathit{e_i}$  probability of
belonging to the structure of  interest $\mathit{p(y_i = 1|\bx_i)}$ or to
the background, $\mathit{p(y_i = 0|\bx_i)}$.

The  optimal subgraph  $\bT^*$  can  then  be taken  to  be tree  that
minimizes the cost function over all trees $\bT$ that are subgraphs of $\bG$

\begin{equation}
\sum_{e_i\in E_{\bT}}-\log\frac{p(y_i = 1|\bx_i)}{p(y_i = 0|\bx_i)}  ,
\label{eq:ObjF}
\end{equation}
where $E_{\bT}$ represents the edges of  $\bT$.  Provided that one does not take
into account the geometry  of the tree but only its topology,  this can be shown
to be Maximum a Posteriori estimate.  In practice, however, it is more effective
to formulate the MAP problem in terms  of pairs of consecutive edges. This makes
it  possible to  introduce better geometric  constraints~\cite{Turetken12} and  to find
generic subgraphs as opposed to only trees~\cite{Turetken13a}.

\begin{figure}[t]
\centering
\subfloat[]{\includegraphics[height=0.2\textwidth]{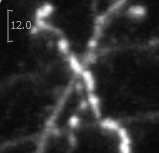}\label{fig:neuron_intersection}}\hspace{1em}
\subfloat[]{\includegraphics[height=0.2\textwidth]{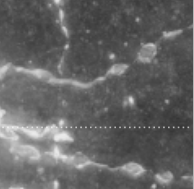}\label{fig:neuron_unevenStaining}}\\

\subfloat[]{\includegraphics[height=0.2\textwidth]{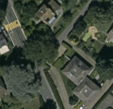}\label{fig:road_occluded}}\hspace{1em}
\subfloat[]{\includegraphics[height=0.2\textwidth]{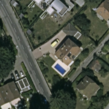}\label{fig:road_driveway}}
\caption{Ambiguous image regions. (a)  Branch intersection.  (b) Discontinuities
  due to  uneven tissue  staining.  (c)  Discontinuities due  to occlusion  by a
  tree.  (d) Linear structures such as driveways that should be ignored.}
\label{fig:ambiguousCases}
\end{figure}

Whether  using single  edges or  pairs,  the key  requirement for  this kind  of
approach  to  perform  well  is  that  the  classifiers  used  to  estimate  the
probabilities of Eq.~\ref{eq:ObjF} should be well-trained. This is especially important
in ambiguous parts of the images such as those depicted by Fig.~\ref{fig:ambiguousCases}.

This  necessitates
significant amounts of ground-truth annotations to capture the large variability
of the  data and  to cope  with imaging  artefacts and  noise. To decrease the amounts of necessary time and  effort, we  introduce an  AL algorithm  that is
suited to  delineation problems represented  on a  graph.  At each  iteration it
selects a sequence of consecutive edges  from an overcomplete graph, such as the
one  described above,  which should  be labeled  next in  order to  decrease the
uncertainty in the most ambiguous image regions.

In theory the sequences  could be of arbitrary length, that is  1,2, or more. In
practice, we  will see that  2 is near optimal  because 2 consecutive  edges are
enough to capture some amount of geometry and because querying at each iteration more than 2 edges does not update the model frequently enough.

In the results  section, we will use the  algorithm of~\cite{Turetken13a}, which
operates  on edge  pairs to  produce the  final
delineations.\footnote{The code is  not publicly available but  the authors gave
  us a binary version of it.} However, our approach is generic and could be used
in conjunction  with any delineation pipeline  that represents the problem  on a
graph and requires supervised edge classification.

\section{Approach}
\label{sec:approach}

In this  section, we  first cast the  traditional Uncertainty Sampling approach
into  our chosen  delineation  framework.   We then  introduce  our approach  to
probability propagation designed to rapidly identify ambiguous image regions and prevent the so-called sampling bias that may lead the classifier to  explore irrelevant parts of the feature space.
Finally, we combine this with an approach to batch density-based learning that simplifies the
interaction while  guaranteeing that  the batches  are representative and diverse enough to
achieve rapid convergence.

\subsection{Random and Uncertainty Sampling}
\label{sec:uncertainty}

The simplest strategy for picking samples  to be annotated is to randomly choose
them from a pool of unlabeled ones in so called Random Sampling (\textbf{RS}). 
As discussed in  Section~\ref{sec:related}, Uncertainty Sampling (\textbf{US}) is a
simple and popular approach to more efficient learning by querying first the
most uncertain samples according to a metric, such as Shannon entropy.

In our  case, as discussed  in Section~\ref{sec:delin},  each edge $e_i$  of the
spatial  graph $\bG$  is assigned  a feature  vector $\bx_i$  computed from  the
pixels surrounding the corresponding path.  Let
\begin{equation}
  p_{t}(y_i=y|\bx_i) \mbox{ for } y \in \{0,1\}
  \label{eq:activeP}
\end{equation}
be the  probabilities computed by  classifier $C_t$  after $t$ AL  iterations that
$e_i$ lies on the centreline of a true structure or not. Let also $S_t$ be the set
of  $N_t$ annotated  samples $(\bx_j,y_j)_{1  \leq j  \leq N_t}$  used to  train
$C_t$.  $p_0$ denotes  the probabilities  returned by  the classifier  using the
small initial batch $S_0$ of annotated  samples. When training is complete after
$T$ iterations, $p_T$  is then used to compute the  probabilities that appear in
Eq.~\ref{eq:ObjF}.  

Given a classifier  $C_{t-1}$ trained  using  the  training  set  $S_{t-1}$,  AL
iteration $t$ involves choosing one or  more unlabeled edges, asking the user to
label them, adding them to the training set $S_{t-1}$ to form $S_{t}$
and, finally, training classifier $C_{t}$.  In \RS{}, this is done by
randomly picking one or  more $\bx$  not already  in $S_{t-1}$.  In \US{}, it is done by computing for each $\bx$ the entropy:
\begin{eqnarray}
  H(\bx) & = & - \log(p_{t-1}(y=0|\bx))p_{t-1}(y=0|\bx) \nonumber \\
         & & - \log(p_{t-1}(y=1|\bx)) p_{t-1}(y=1|\bx) \; \; 
\label{eq:entropy}
\end{eqnarray}
and selecting  the vector(s)  with the highest entropy.   Since $H(\bx)$  is largest
when the  classifier returns a  $0.5$ value and  minimum when it  returns values
close to zero or one, this assumes that those vectors whose probability of being
a true path  is computed to be $0.5$  are the most uncertain and  closest to the
decision boundary. Therefore annotating them is  likely to help refine the shape
of that boundary.

This approach  can be  effective but it  can also fall victim to  {\it sampling
  bias}. This  happens when  the current  classifier is  so inaccurate  that its
decision boundary is far away from the real one and the learner ends up focusing
on an irrelevant  part of the feature  space. Our approach is  designed to avoid
this trap. 


\subsection{Probability Propagation}
\label{sec:propagation}

The probability  $p_t$ returned by  the path  classifier takes into  account the
appearance of  only a  single path.   By doing so,  it neglects  the information
present in  the wider neighborhood, provided  by the
other paths  in the  graph that  share an  endpoint with  it. In  particular, it
ignores the fact that contiguous paths are  more likely to share labels than non
contiguous ones.


To  account for this,  we  took inspiration from the semi-supervised learning method of~\cite{Zhou04} and
implemented a modified  version of it  that propagates probabilities
instead of labels. There, the label propagation is used to classify a large pool of unlabelled examples having only a few labelled instances. In our Probability Propagation Sampling ($\PPS$) strategy we propagate the probabilities assigned by the base classifier to identify samples that differ significantly from their neighbourhood. 

Let $\bP_0$ be an  $N \times 2$ matrix.  Its entries are the
probabilities $p_{t}(y_i=y|\bx_i)$  of Eq.~\ref{eq:activeP} for all  $N$ samples
and $y \in  \{0,1\}$, except for already  annotated ones for which  we clamp the
values  to zero  or  one depending  on  their label.   The  information is  then
propagated as follows:
\begin{enumerate}
  
\item   Build  an   $N   \times   N$  affinity   matrix   $\bW \in \mathbb{R}^{N \times N}$  with   elements
  $w_{ij}=\exp(-||\mathbf{x_i} - \mathbf{x_j}||^2/2\sigma^2)$ if $e_i$ and $e_j$
  share a node and zero otherwise.
  
\item Build  a symmetric matrix  $\bS = \bD^{-1/2}\bW\bD^{-1/2}$,  where 
  $\bD$ is diagonal with elements $d_{ii}~=~\sum_j w_{ij}$.
  
\item  Iterate   $\bP^{i+1}  =  \alpha  \bS\bP^i+(1-\alpha)\bP_0$   followed  by
  normalization   of  the   rows   of  $\bP^{i+1}$   until  convergence,   where
  $\alpha\in(0,1)$ specifies  how much  information is exchanged  between neighbors
  and how much of the original  information is retained. The series was shown to converge to   $\bP^*  = (\bI-\alpha\bS)^{-1} \bP_0$~\cite{Zhou04} and we will use the closed-form solution.
\end{enumerate}

After the probability propagation, we can compute the entropy of each path at AL iteration \textit{t}, but this time using the new estimates of probability $\mathit{p^*}$:

\begin{eqnarray}
  H(\bx) & = & - \log(p^*_{t-1}(y=0|\bx)) p^*_{t-1}(y=0|\bx)  \nonumber \\
         & & - \log(p^*_{t-1}(y=1|\bx)) p^*_{t-1}(y=1|\bx) \; \; 
\label{eq:entropyPropagated}
\end{eqnarray}

\subsection{Density-based Batch Query}
\label{sec:batch}

The scheme  of Section~\ref{sec:propagation} involves retraining  the classifier
each time  the user  has annotated  a new sample,  forcing them to wait  for the
computation   to  be   over   before  intervening   again.    As  discussed   in
Section~\ref{sec:related}, this is impractical  and most practical AL approaches
work in  batch-mode, that is,  they allow the  user to annotate  several samples
before retraining.

In our case, samples are image  paths and it is much  easier to sample
several paths in the  same image region than over a wide space, which would
imply  scrolling through  a  potentially  large 2D  image  or,  worse, 3D  image
stack. Our solution  to this is to present the  annotator with consecutive paths
represented   by    adjacent   edges   in    the   spatial   graph    $\bG$   of
Section~\ref{sec:delin}. However, in order to be effective, individual paths should be:
\begin{enumerate}
  
\item  {\it  informative}  to  ensure  that  the  new  labels  truly  bring  new
  information,
  
\item {\it representative}, that is, inliers of the statistical distribution
  of all samples,
  
\item {\it  diverse}, that is,  different from each  other and from  the already
  labeled ones.

\end{enumerate}
The entropy measure of Eq.~\ref{eq:entropyPropagated}  can be used to assess the
first of  these three desirable  properties. To measure  the other two, we use the
$N  \times N$ affinity matrix  $\mathbf{\widetilde{W}}$  obtained using the same parameters as the matrix $\bW$ of Section~\ref{sec:propagation}, but whose
elements are measures of pairwise similarity  between \textit{each} of the $N$ samples in the
feature space, not only the neighbours in image.

Let $\bL$ be the indices of already labelled edges and $\bE_k$ be the set of
all possible edge index combinations denoting $k$ consecutive paths. For each $E
\in E_k$,
we can compute the following similarity measures:
\begin{eqnarray}
\sigma_G(E) & = & \sum_{i \in E} \sum_{1 \leq j \leq N} w_{ij} \\
\sigma_L(E) & = & \sum_{i \in E} \sum_{l \in L} w_{il} \\
\sigma_I(E) & = & \sum_{i \in E} \sum_{j \in E, j\neq i} w_{ij},
\end{eqnarray}
where $\sigma_G(E)$  is  a  global  similarity measure, $\sigma_L(E)$  measures
similarity to  already labelled samples  and $\sigma_I(E)$ similarity  within the
batch.  Intuitively, we want to maximize $\sigma_G$ to ensure representativeness
and minimize $\sigma_L$  and $\sigma_I$ to improve diversity and explore the whole feature space.  We therefore take
\begin{equation}
  \mu(E) = \frac{\sigma_G(E) - \sigma_L(E) - \sigma_I(E)}{\sigma_G(E)},
  \label{eq:diversity}
\end{equation}
to be our measure of both  diversity and representativeness.  This formulation does not require any additional parameters to weigh the effect of the three conditions or constructing any additional graphs in the feature space. 

\subsection{Combining Informativeness and Density Measure}
\label{sec:combining}
$\PPS$ allows us for taking advantage of the current model while density-based query enables exploration of the feature space. In order to combine those two effects at each AL iteration we query the batch 
\begin{equation}
E^* = \argmax_{E \in \bE_k} \mu(E) (\sum_{i \in E} H(\bx_i)) \;
\label{eq:combined} 
\end{equation}
where         $H$        is         the        entropy         measure        of
Eq.~\ref{eq:entropyPropagated} and $\mu(E)$ is calculated as in Eq.~\ref{eq:diversity}. In our Density-Probability Propagation Sampling ($\DPS$) the effects of exploration and exploitation are balanced during AL.  

\section{Results}
\label{sec:results}

\begin{figure*}[t]
\centering
\subfloat[]{\includegraphics[width=0.25\textwidth]{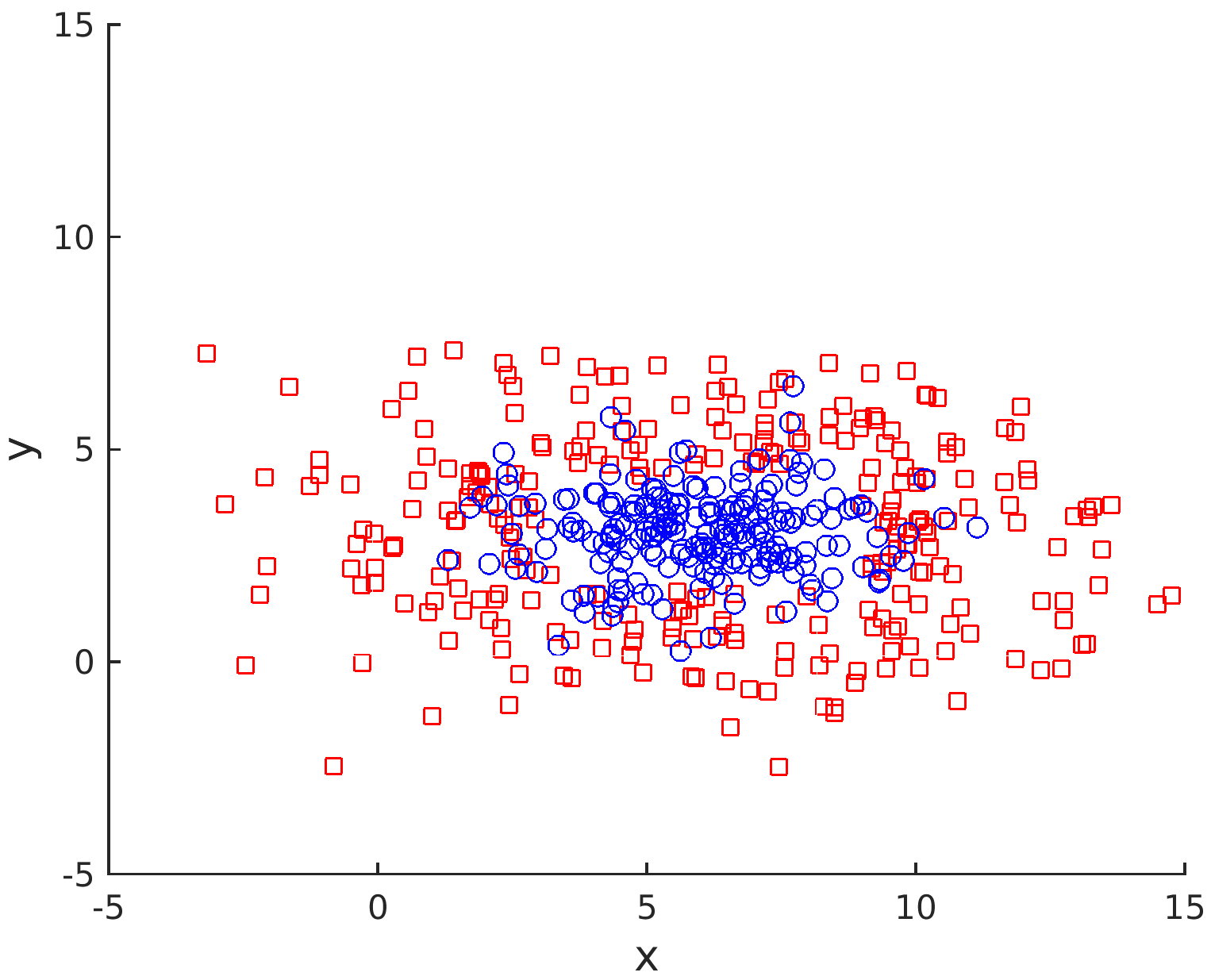}\label{fig:imageSpace}}
\subfloat[]{\includegraphics[width=0.25\textwidth]{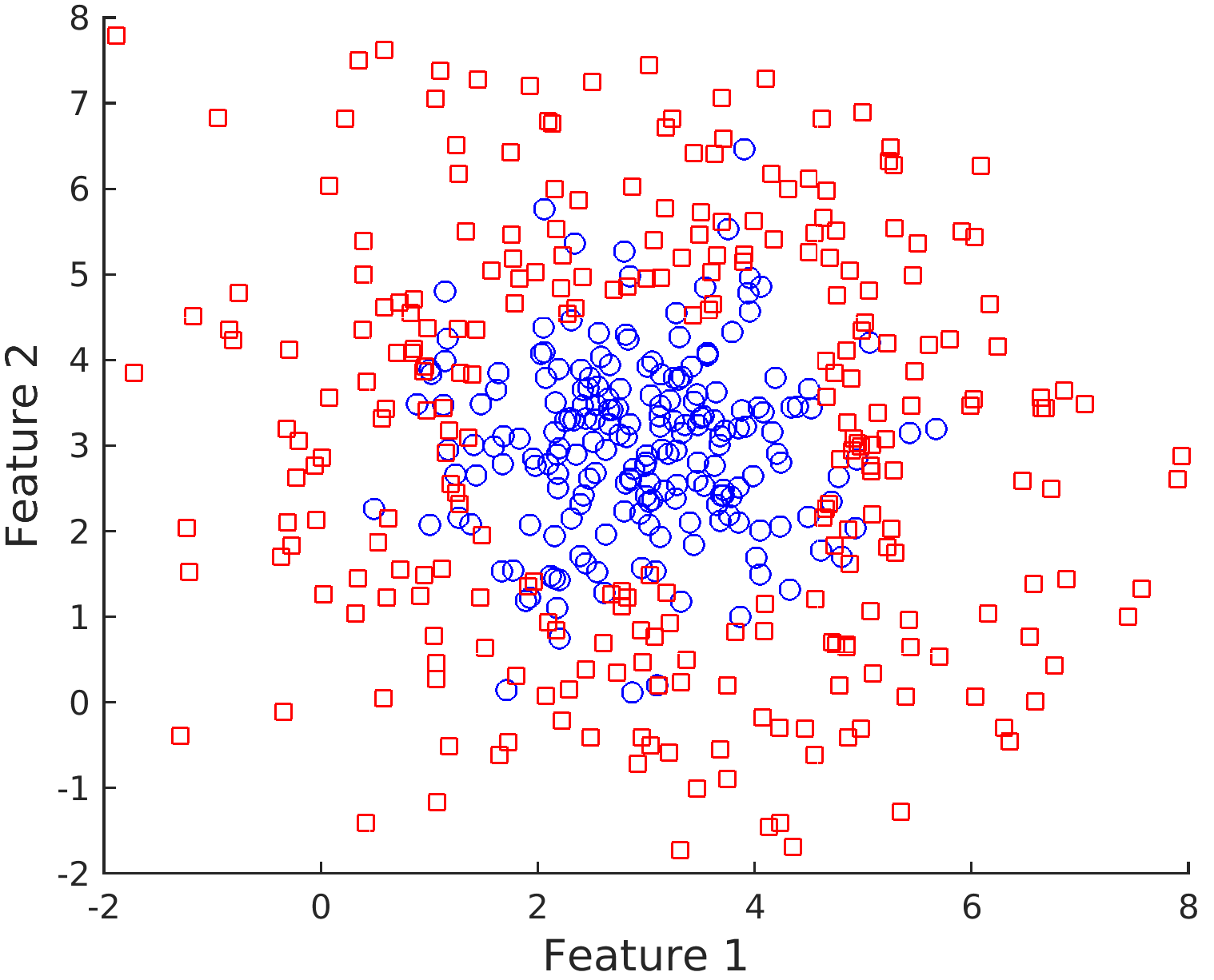}\label{fig:featureSpace}}
\subfloat[]{\includegraphics[width=0.25\textwidth]{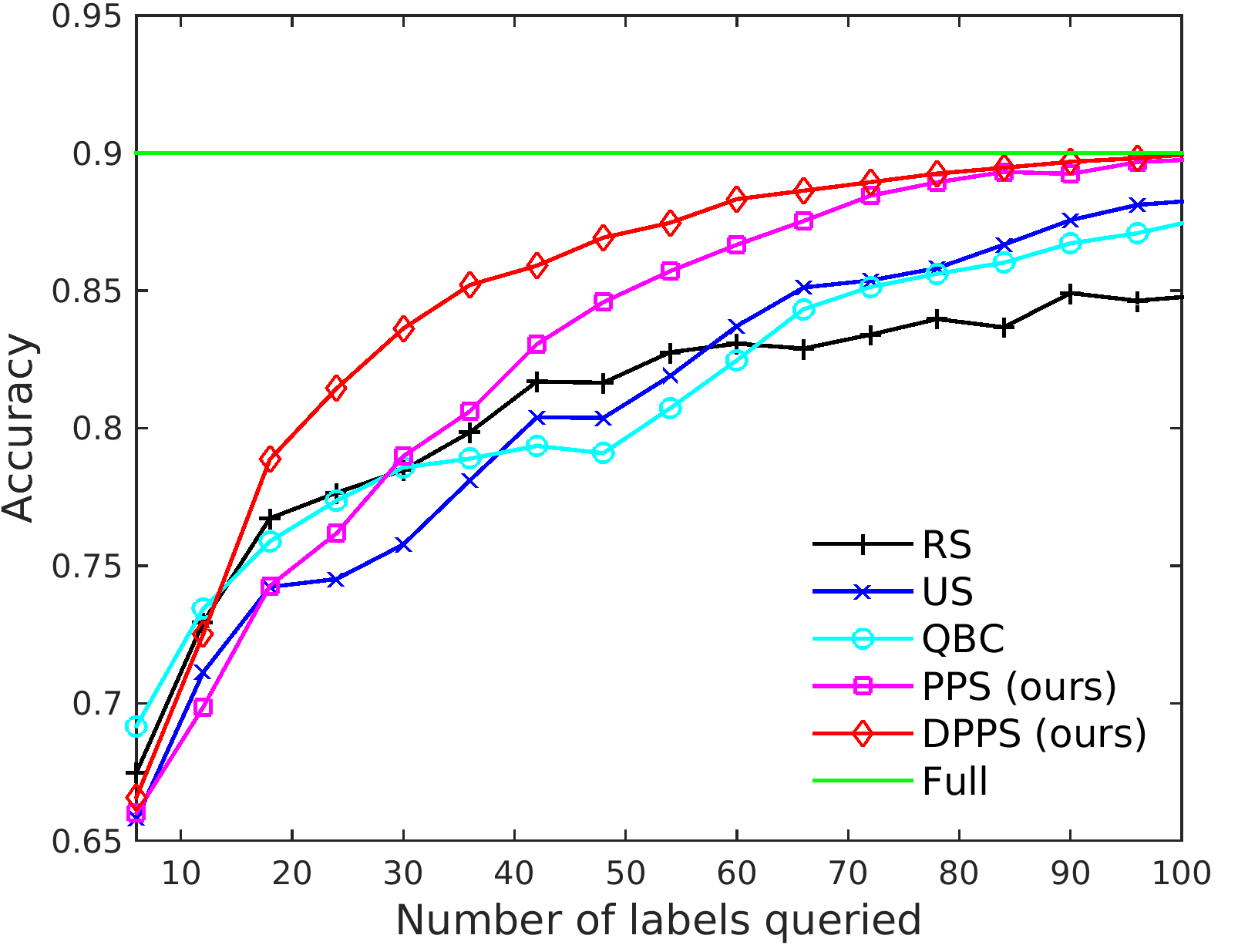}\label{fig:syntheticResults}}\\

\subfloat[]{\includegraphics[width=0.95\textwidth]{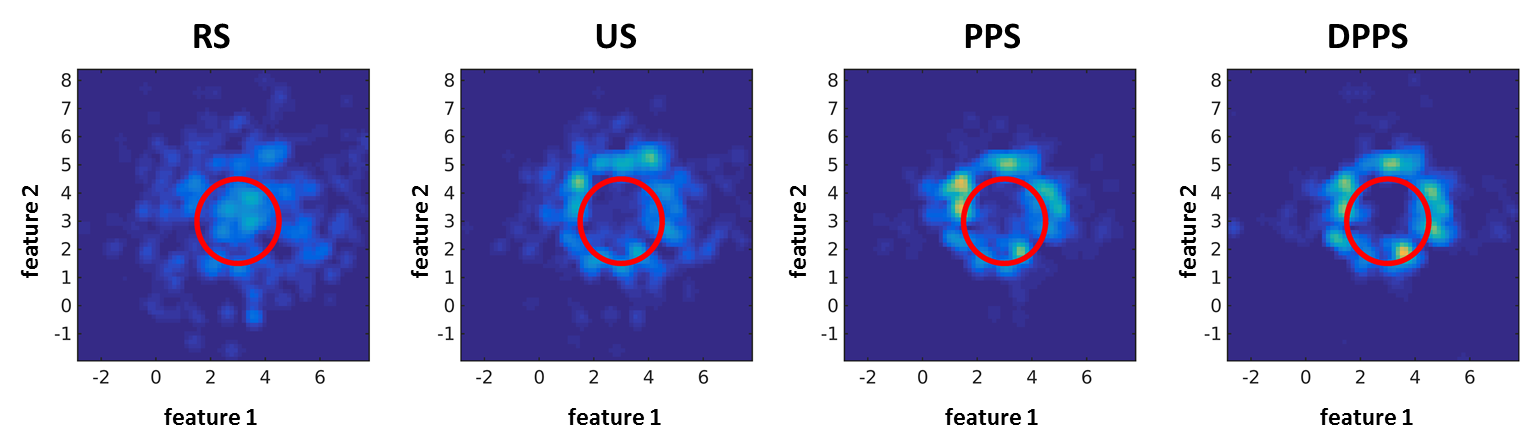}\label{fig:syntheticRandom}}      
\caption{Synthetic dataset: (a) samples in the Euclidean space (b) samples in the feature space.  (c) Classification results. (d) Query heat-maps in the feature space; the red circle indicates the optimal decision boundary. Best viewed in color.}
\label{fig:syntheticExample}
\end{figure*}

In this  section, we  present our  results; we  first describe  our experimental
setup and baselines. We then introduce a synthetic dataset to help visualize the
query decisions  made by  the different  strategies. Finally,  we show  that our
approach outperforms the conventional techniques on four real datasets.

\subsection{Experimental Setup}
\label{sec:setup}

We apply  our AL approach for  reconstruction of curvilinear networks  in 2- and
3-D images. As discussed in Section~\ref{sec:delin},  the overcomplete
  graphs, as well as the final delineations obtained once the classifiers have been properly
  trained are constructed using the delineation algorithm of~\cite{Turetken13a}. The feature  vectors associated to each path are  based on Histogram
  of Oriented Gradients  specially designed for linear  structures. They capture
  the contrast, orientation, and symmetry of the paths.

  The probabilities  of Eq.~\ref{eq:ObjF}  are computed  by feeding  the feature
  vectors to Gradient Boosted Decision Trees~\cite{Becker13b} with
  an  exponential  loss. We  found it  be well  suited to  interactive
  applications because it can  be retrained fast, that is in under  3s for all the
  examples we show in this paper. To avoid overfitting especially in the initial
  stages of AL, we set the number of  weak learners to 50, maximum tree depth to
  2  and shrinkage  to  0.06. Each  tree  is optimized  using  50\% of  randomly
  selected  data. Out  of possible  303 features,  50 are  investigated at  each
  split. The classifier  returns score \textit{F} that can be  then converted to
  probability using the logistic correction~\cite{Niculescu05}, that is,
\begin{equation}
p(y=1|x) = \frac{1}{1+\exp(-2F(x))} \; .
\end{equation}
The edge connectivity matrix of Section~\ref{sec:propagation} is computed on the
basis of the overcomplete graphs.

The annotated  ground truth data we  have for all  datasets, allows us to
simulate the user  intervention. We assume edges that are 10 pixels/voxels apart from the corresponding ground-truth path and with a normalised intersection exceeding 0.5 to be positive.  We start each query by a  random selection of 4 data points belonging
to  each  class  (background/network).   Unless stated  otherwise,  we  query  2
consecutive  paths  during  each  iteration  and this  choice  is  explained  in
Section~\ref{sec:real}. We proceed  until the total number  of labelled samples
reaches  100. Each  AL trial  is  repeated 30  times  and the  results are  then
averaged.


\subsection{Baselines}
\label{sec:baselines}

We compare  the  two  versions of  our  approach, Probability  Propagation
Sampling  (\PPS{})  and Density  Probability  Propagation  Sampling (\DPS{})  as
described in Sections~\ref{sec:propagation}  and~\ref{sec:combining}, to
the following baselines:
\begin{itemize}

\item Random Sampling (\RS{}) - selecting a random pair at each iteration.
  
\item Uncertainty Sampling (\US{}) - selecting a pair with the highest sum of individual entropies as given by Eq.~\ref{eq:entropy}.
\item Query-By-Committee (\QBC{}) - selecting a pair that causes the greatest disagreement in a set of hypotheses, here represented by trees in a Random Forest. We measure the disagreement using the definition of~\cite{Dagan95}. 

\end{itemize}
For calibration  purposes, we also  report the  classification performance
using all the available training data  at once (\textbf{Full}), that is, without
any AL.

\subsection{Synthetic Dataset}
\label{sec:synthetic}

To  compare  the qualitative  behavior  of  different  strategies, we  create  a
synthetic dataset.  In the image  space depicted by Fig.~\ref{fig:imageSpace}, a
positive class  is surrounded by  a negative  one, which resembles  what happens
when trying to  find real linear paths surrounded by  spurious ones.  We created
feature space depicted by  Fig.~\ref{fig:featureSpace} by transforming the image
coordinates and  adding random noise  so that  the decision boundary  in feature
space does  not correspond  to the one in Euclidean space.  We built  the required
spatial graph  by connecting  each point  to its  10 nearest-neighbors  in image
space. We compute the weighting matrix $\bW$ using RBF kernel with $\sigma$ = 1 and set probability propagation $\alpha$ to 0.9.

As can be  seen in Fig.~\ref{fig:syntheticResults}, \PPS{}  and \DPS{} outperform
the baselines and  after querying 90 examples match the  performance obtained by
training on  the whole training  set.  This corresponds  to a 80\%  reduction in
annotation effort. Furthermore,  \DPS{} does better than \PPS{}  early on, Later
in  the process,  once  the  most informative  examples are all labeled,  both
approaches yield similar results.

In Fig.~\ref{fig:syntheticRandom},  we use  a heat  map in feature space to  depict the  the most
frequently queried  regions and  overlay the optimal  decision boundary  in red.
They indicate that  propagating information in a spatial graph  helps refine the
feature  space  faster  than  simple  uncertainty  query.   Introducing  density
measures further constrains  the search space making the  process more effective
and  sampling  more uniformly  around the optimal decision boundary. 

\begin{figure*}[t]
\centering
\subfloat[]{\includegraphics[height=0.2\textwidth]{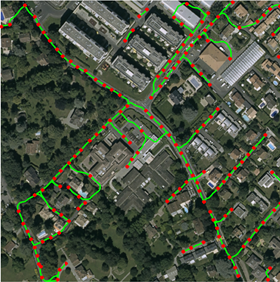}\label{fig:roadsDouble}}\hspace{0.5em}
\subfloat[]{\includegraphics[height=0.2\textwidth]{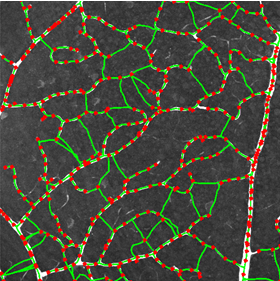}\label{fig:BVDouble}}\hspace{0.5em}
\subfloat[]{\includegraphics[height=0.2\textwidth]{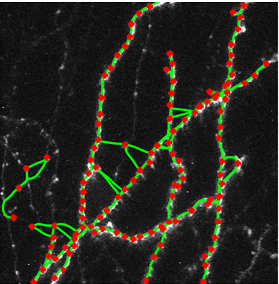}\label{fig:neuronsDouble}}\hspace{0.5em}
\subfloat[]{\includegraphics[height=0.2\textwidth]{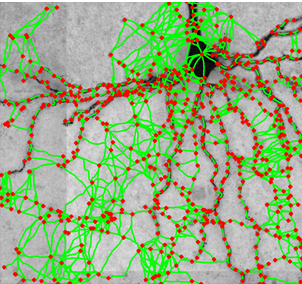}\label{fig:BFDouble}}
\caption{Training images with superimposed overcomplete graphs (a) \textit{Roads} (b)  \textit{Blood vessels} (c) \textit{Axons} (d) \textit{Brightfield neurons}.}
\vspace{-5mm} 
\end{figure*}

\begin{figure*}[t]
\centering
\subfloat[]{\includegraphics[height=0.2\textwidth]{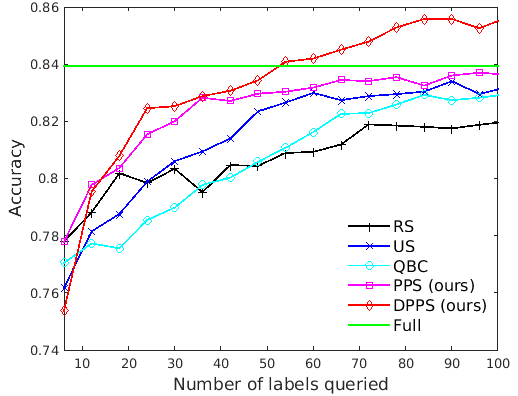}\label{fig:resultsR}}
\subfloat[]{\includegraphics[height=0.2\textwidth]{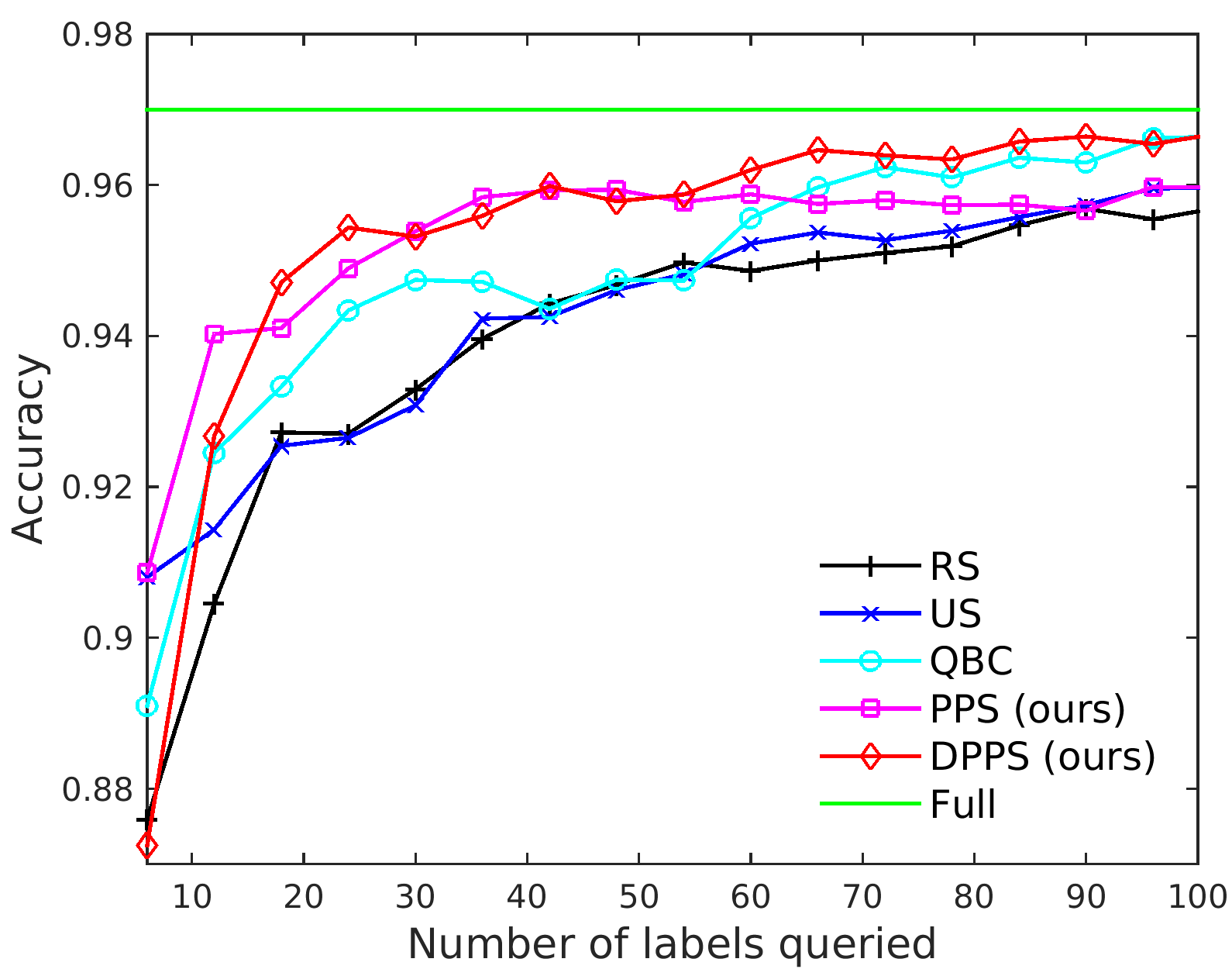}\label{fig:resultsBV}}
\subfloat[]{\includegraphics[height=0.2\textwidth]{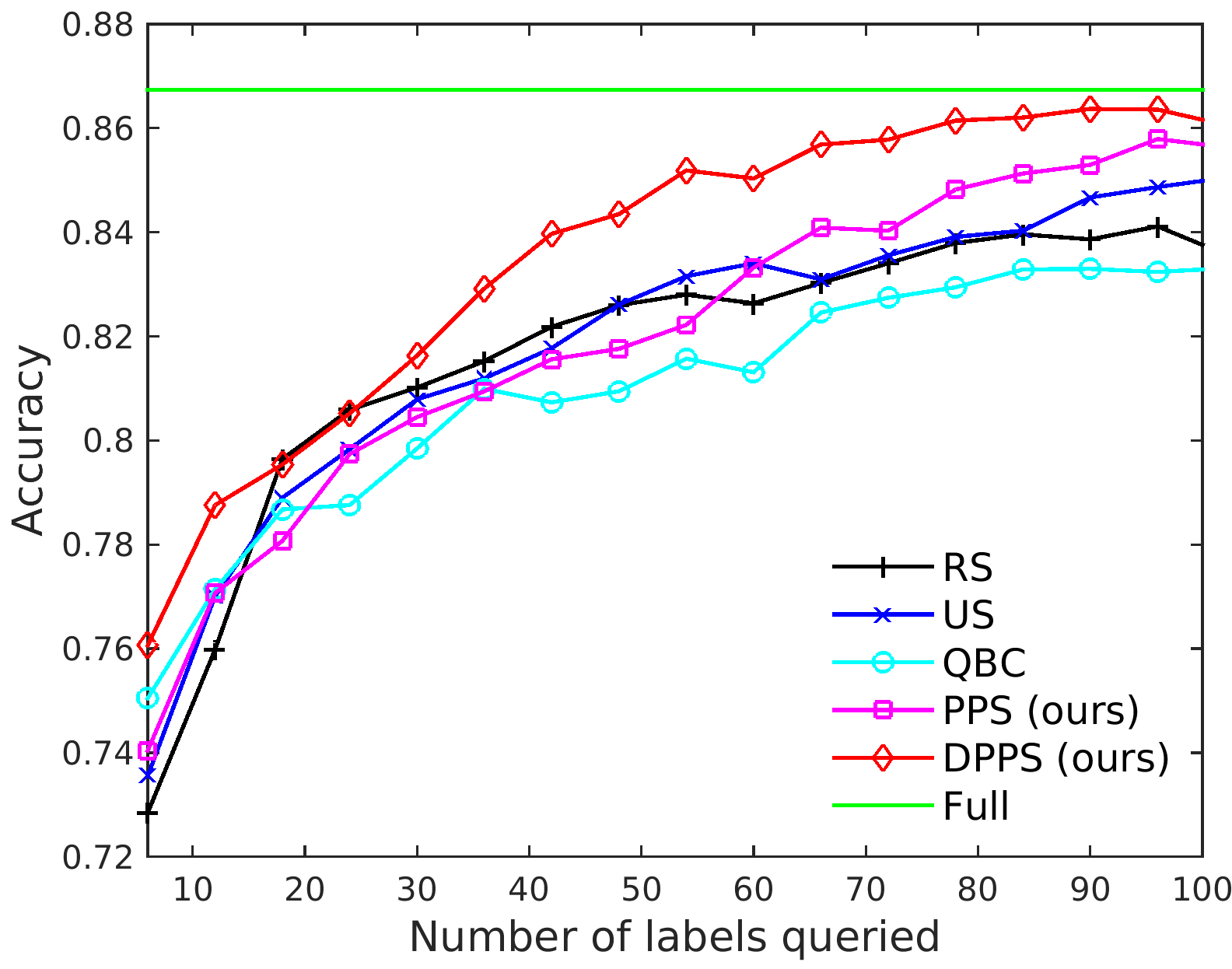}\label{fig:resultsN}}
\subfloat[]{\includegraphics[height=0.2\textwidth]{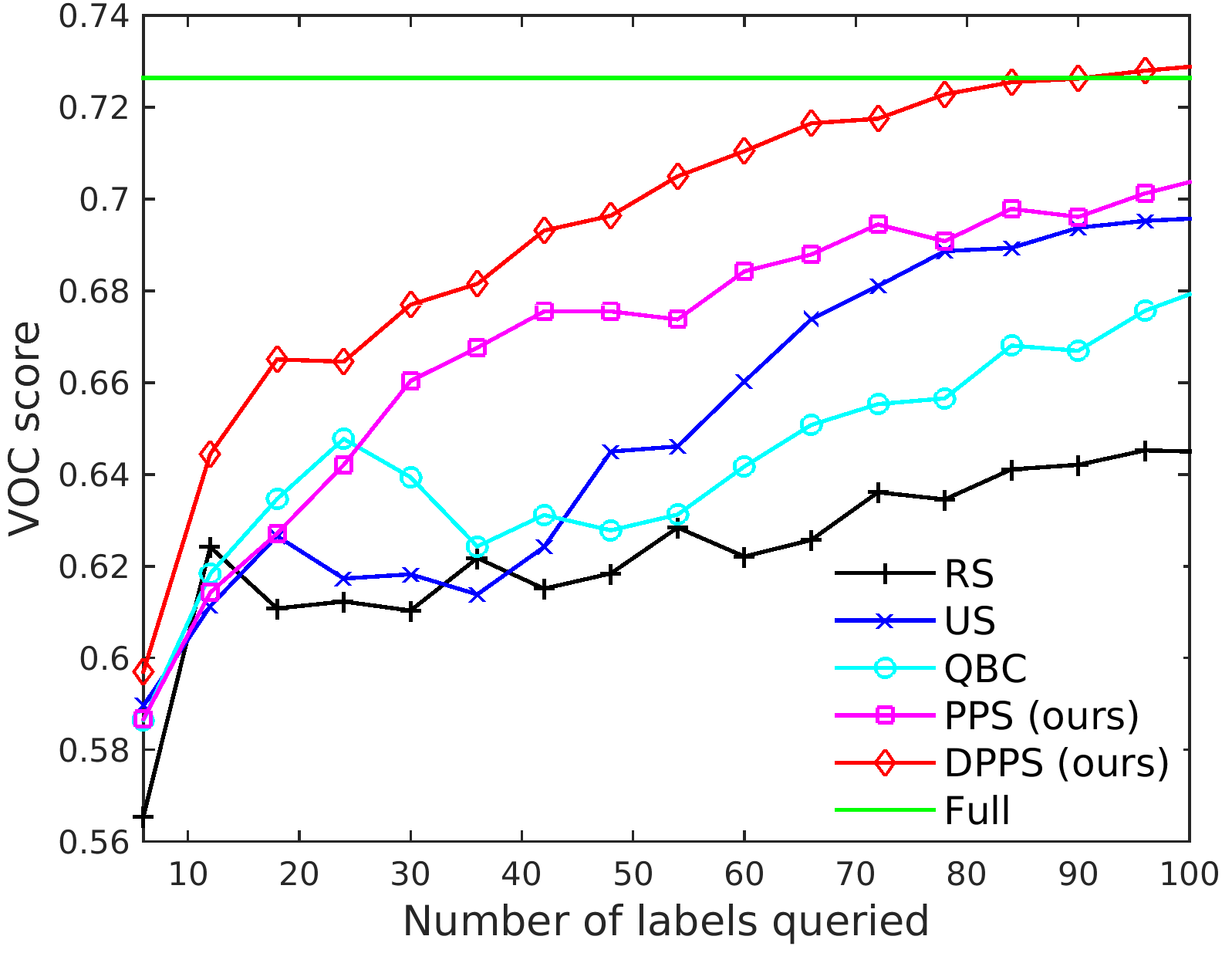}\label{fig:resultsBF}}
\caption{Classification results for the (a) \textit{Roads}  (b) \textit{Blood vessels} (c) \textit{Axons} (d) \textit{Brightfield neurons} datasets.}
\vspace{-3mm}
\end{figure*}

\subsection{Real Datasets}
\label{sec:real}
\paragraph{Roads}

The dataset  consists of 2D  aerial images of  roads. They  include road
patches occluded by trees and contain road-like structures such as driveways, thus making the classification task difficult.

\begin{figure*}[t]
\centering
\includegraphics[width=0.97\textwidth]{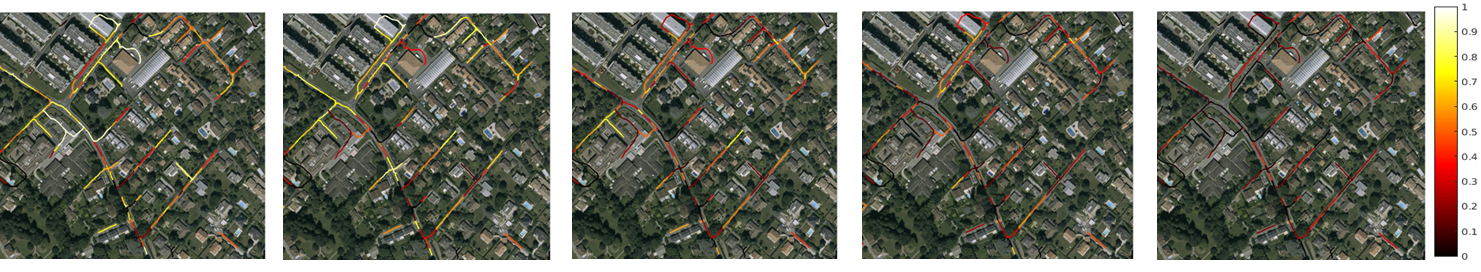}
\caption{Color-coded ambiguity of the training paths given by Eq.~\ref{eq:combined} after 5, 15, 25, 35 and 45 AL iterations. The measure was normalized across the iterations. There is a clear decrease in the uncertainty as the learning progresses. Best viewed in color.}
\label{fig:progress}
\vspace{-5mm}
\end{figure*}

\begin{figure*}[t]
\centering
\subfloat[]{\includegraphics[height=0.21\textwidth]{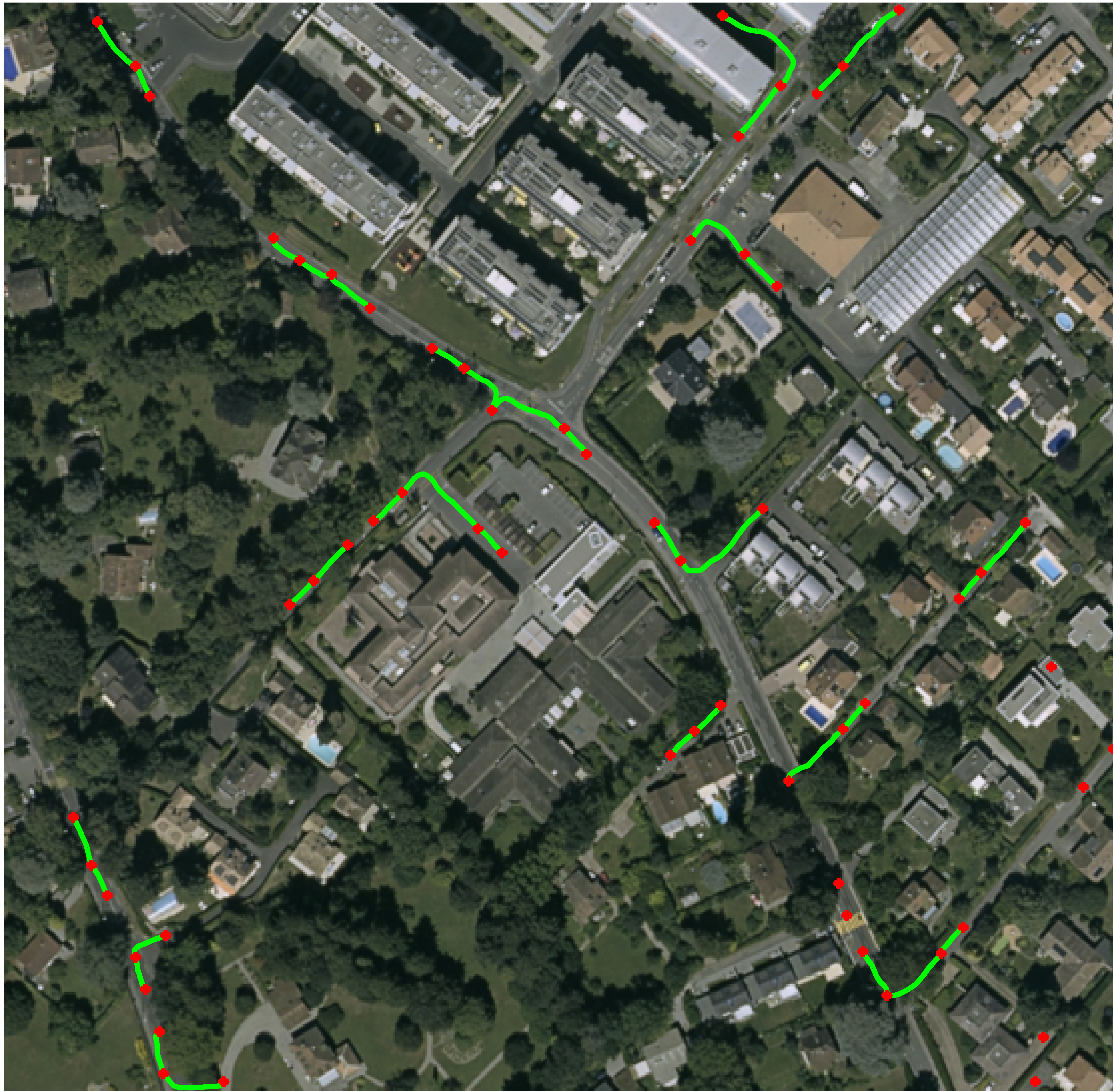}\label{fig:uncertainPairsR}}\hspace{0.1em}
\subfloat[]{\includegraphics[height=0.21\textwidth]{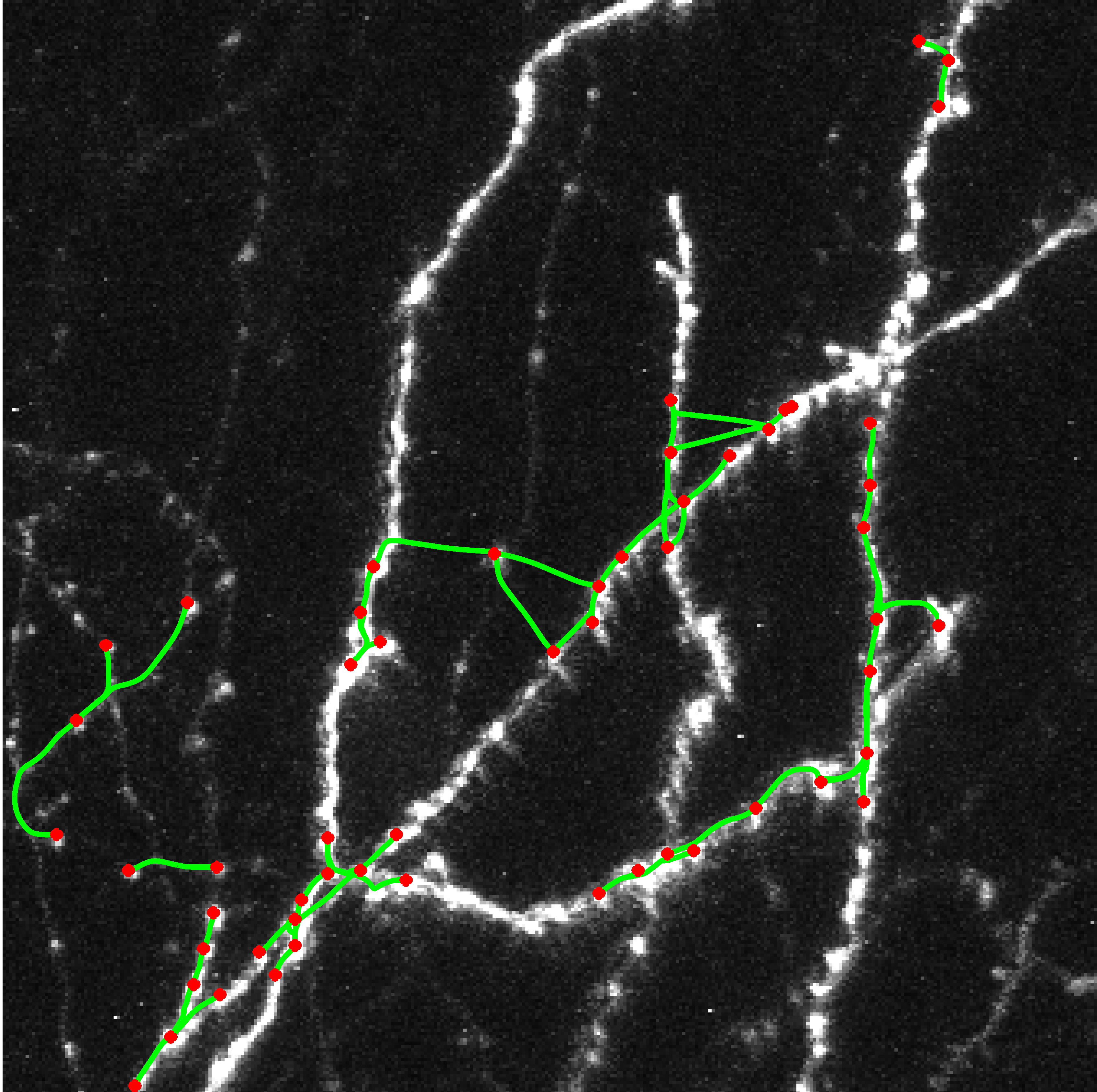}\label{fig:uncertainPairsN}}\hspace{0.2em}
\subfloat[]{\includegraphics[height=0.21\textwidth]{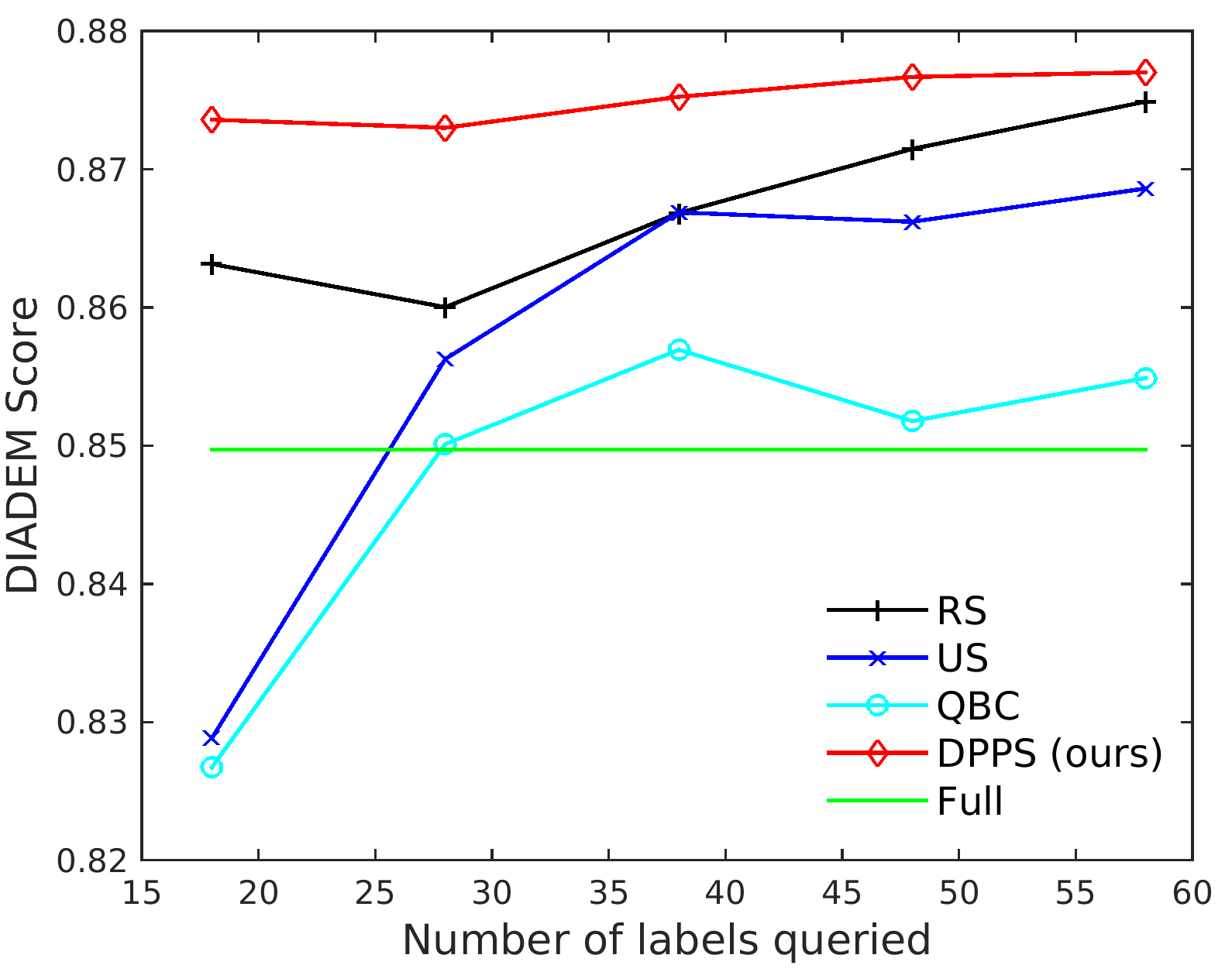}\label{fig:DIADEMRoads}}
\subfloat[]{\includegraphics[height=0.21\textwidth]{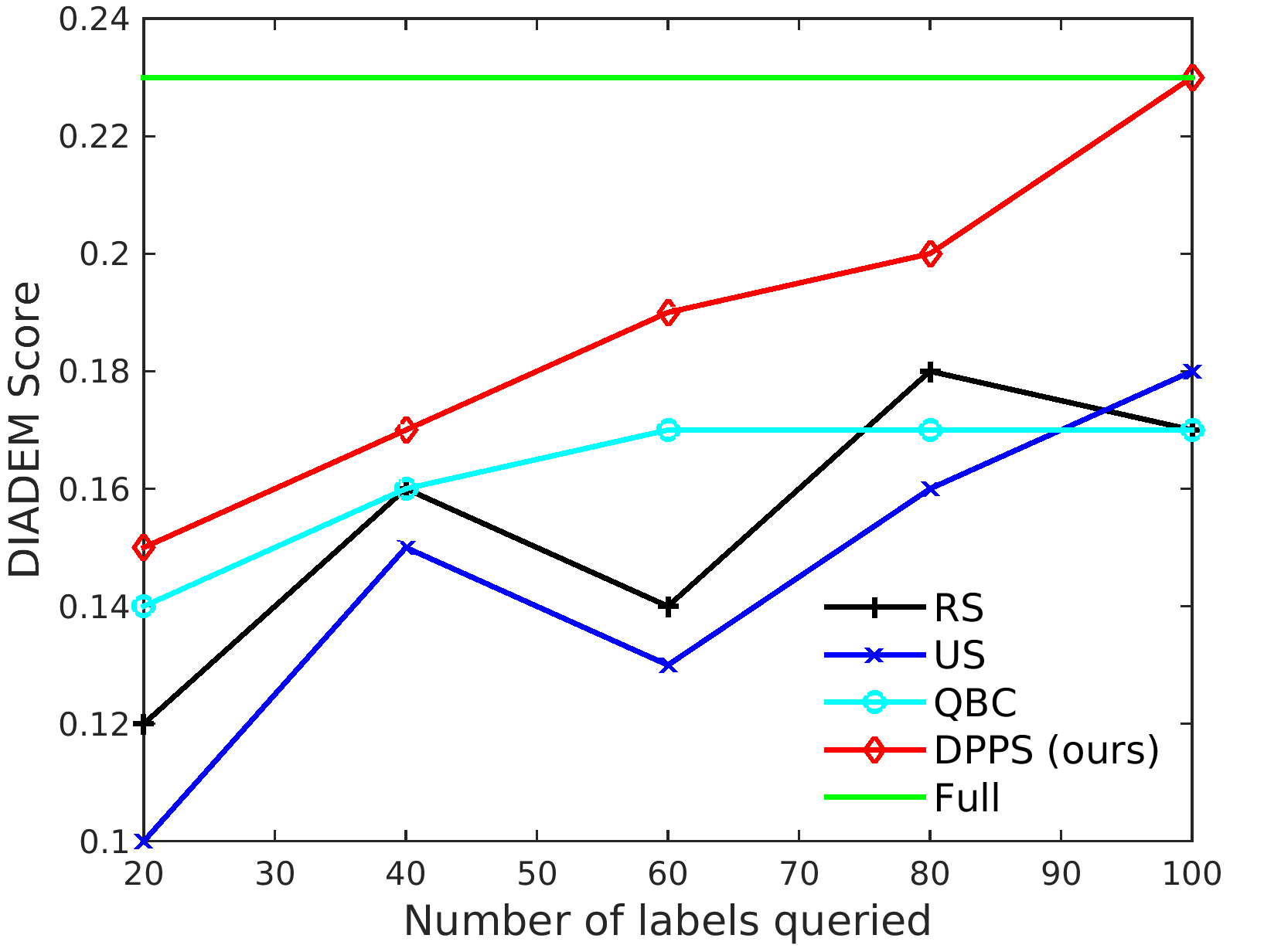}\label{fig:DIADEMAxons}}
\caption{The most frequently queried samples for  the (a) \textit{Roads} and (b) \textit{Axons} datasets. They often coincide with the ambiguous cases as discussed in Section~\ref{sec:delin}. Averaged DIADEM scores of final reconstruction for the (c) \textit{Roads} and (d) \textit{Axons} datasets.}
\vspace{-5mm}
\end{figure*}

We compute the weighting matrix $\bW$ using RBF kernel with $\sigma$ = 1 and set probability propagation $\alpha$ to 0.9. As shown in Fig.~\ref{fig:resultsR}, both our approaches outperform the baselines
and reach the full-dataset performance after as few as 50 samples, which corresponds
to  75\%  reduction in  annotation  effort. Interestingly, the  accuracy  keeps
increasing above the  \textbf{Full} dataset accuracy. This  behavior was already
reported in~\cite{Schohn00} and suggests that in some cases a well chosen subset
of data  produces better generalization than  the complete set. In Fig.~\ref{fig:progress} we present how the uncertainty about the training paths (as measured by Eq.~\ref{eq:combined}) decreases as AL progresses. The  analysis of
the  most frequently  queried  samples  shown in  Fig.~\ref{fig:uncertainPairsR}
reveals  that our  method selects  mainly the  occluded paths  and those at the
intersections between  two roads of different  sizes or a road  and a driveway. They
correspond  to  the ambiguous  cases  discussed  in Section~\ref{sec:delin}  and
presented in Fig.~\ref{fig:road_occluded} and Fig.~\ref{fig:road_driveway}. This makes it possible to learn the correct connectivity pattern and avoid mistakes as we postulated in Section~\ref{sec:delin}. To verify this, we compare not only the classification performance, but also the quality of the final reconstruction. We run the full reconstruction framework with
classification followed by an optimization step and evaluate the reconstruction using the DIADEM score~\cite{Ascoli10}. It ranges from 0 to 1 with 1 being a perfect reconstruction. As shown in Fig.~\ref{fig:DIADEMRoads}, our approach outperforms the baselines also in terms of the quality of the final reconstruction. Interestingly, we again get a better result than by training with the \textbf{Full} dataset.

These results  were obtained  by querying  pairs of  edges. To  test the
  influence of the length of the sequences  we query, as discussed at the end of
  Section~\ref{sec:delin}, we reran the experiments using singletons, pairs, and
  triplets. As can  be seen in Fig.~\ref{fig:longQuery} and in the supplementary material for the other datasets, using pairs tends to give the best results and   this  is   what  we   will  do  in   the  remainder   of  this
  paper. Note that we assume that annotating one edge counts as one label, but in reality the effort of annotating several \textit{consecutive} edges is less than labeling the same number of instances at random locations, as the user does not need to scroll from one region to another.

\paragraph{Blood vessels}
The image stacks depicting direction-selective retinal ganglion cells were acquired with confocal microscopes. They contain many cycles and branch crossings. We compute the weighting matrix $\bW$ using RBF kernel with $\sigma$ = 0.7 and set $\alpha$ to 0.9. As shown in Fig.~\ref{fig:resultsBV}, our two methods bring about considerable improvements, especially at the beginning of AL and in the second part perform similarly to \QBC{}. 

\paragraph{Axons}
dataset consists of 3D 2-photon microscopy images of axons in a mouse brain. The main challenge associated with these images is low resolution in the z-dimension resulting in some disjoint branches being merged into one, which drastically changes the connectivity of the final solution. 

We compute the weighting matrix $\bW$ using RBF kernel with $\sigma$ = 3 and set $\alpha$ to 0.9. The accuracy plot (Fig.~\ref{fig:resultsN}) reveals that yet again our methods perform better than the baselines, especially in the later stages of learning, and result in a 65\% reduction in the training effort. As seen in Fig~\ref{fig:uncertainPairsN}, the most frequently queried edges are concentrated in the regions where two branches seem to intersect in the xy-plane. In Fig.~\ref{fig:DIADEMAxons} we show that this again improves the quality of the final reconstruction.

\begin{table}
\begin{center}
\resizebox{\columnwidth}{!}{
\begin{tabular}{|l|c|c|c|c|c|}
\hline
 & \RS{} & \US{} & \QBC{} & \PPS{} & \DPS{} \\
\hline
Roads & 0.00055 & 0.00040 & 0.00052 & 0.00036 & \textbf{0.00031} \\
\hline
Blood vessels & 0.00052 & 0.00071 & 0.00028 & \textbf{0.00019} & \textbf{0.00019} \\
\hline
BF neurons & 0.0040 & 0.0017 & 0.0008 & 0.0007 & \textbf{0.0003}\\
\hline
Axons & 0.00060 & 0.00061 & 0.00047 & 0.00048 & \textbf{0.00043}\\ 
\hline
\end{tabular}
}
\end{center}
\caption{Variance of the results.}
\label{tab:variance}
\end{table}

\paragraph{Brightfield neurons}
The dataset consists of 3D images of neurons from biocytin-stained rat brains acquired using brightfield microscopy. As in the \textit{Axons} dataset, the z-resolution is low. The corresponding training graph is much bigger than in the previous 2 cases and consists of more than 3000 edges, most of which are negative. To assess the performance of different methods, we compute the VOC score~\cite{Pascal-voc-2010} instead of accuracy. This is due to the fact that in this dataset around 95\% of the edges are negative and the VOC score does not take into account true negatives. We compute the weighting matrix $\bW$ using RBF kernel with $\sigma$ = 1 and set $\alpha$ to 0.9. As seen in Fig.~\ref{fig:resultsBF}, our methods outperform \RS{}, \US{} and \QBC{}. For \US{} and \QBC{}, we can notice the possible effects of bias trap, when the performance does not change for a few iterations, even though more and more labels are queried.

Note that each of the experiments was repeated 30 times and the results are averaged. In Table~\ref{tab:variance} we present also the variance of the results. In all cases except for one \PPS{} approach shows smaller variance than the baselines and \DPS{} yields even lower variance.

\begin{figure}[h]
\centering
\includegraphics[height=0.25\textwidth]{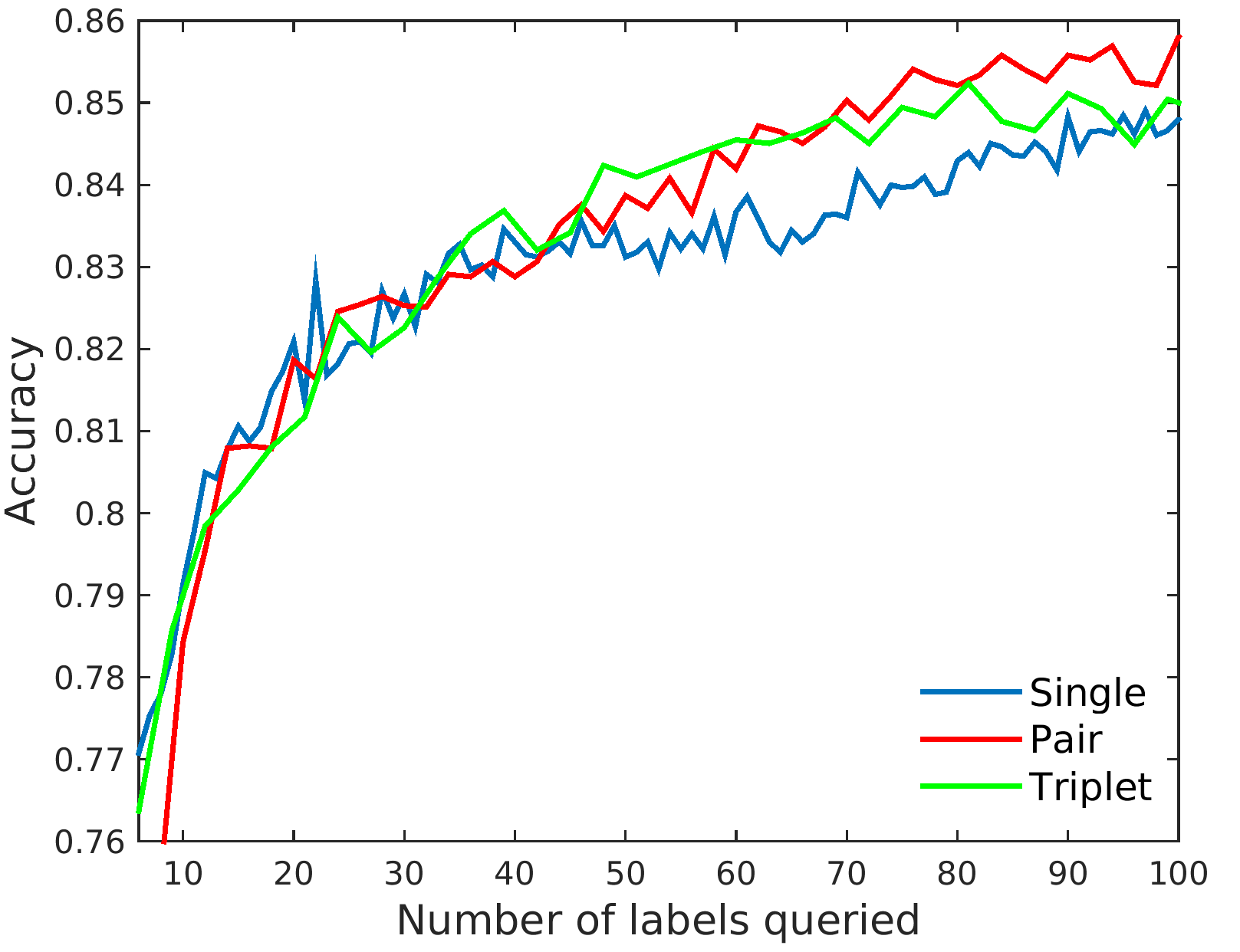}
\caption{The classification performance for different batch sizes for the \textit{Roads} dataset.}
\vspace{-3mm}
\label{fig:longQuery}
\end{figure}


\section{Conclusion}
\label{sec:conclusion}

In this paper we introduced an approach to incorporating the geometrical information that increases the effectiveness of AL for the delineation of curvilinear networks. Additionally, we introduced a density-based strategy, which ensures that the selected batches are informative, diverse and representative of the underlying distribution. It also allows us to query sequences of consecutive paths, further reducing the annotation effort. Our approach showed superior performance for a wide range of networks and imaging modalities when compared to a number of conventional methods.

{\small
\bibliographystyle{ieee}
\bibliography{string,vision,learning,biomed}
}

\end{document}